\documentclass[letterpaper]{article}
\usepackage[T1]{fontenc}
\usepackage{spconf,amsmath,graphicx}
\usepackage[space]{cite}
\usepackage[hidelinks]{hyperref}
\newcommand{\NB}{455}
\newcommand{\NBHl}{229}
\newcommand{\NBPr}{226}

\newcommand{\JudgmentsB}{13,649}

\newcommand{\NBH}{199}
\newcommand{\NBHHl}{99}
\newcommand{\NBHPr}{100}

\newcommand{\JudgmentsBH}{5,868}

\newcommand{\AgeMin}{18}
\newcommand{\AgeMax}{82}
\newcommand{\AgeMean}{42.1}
\newcommand{\AgeSD}{13.1}
\newcommand{\FemaleAllN}{329}
\newcommand{\FemaleAll}{50\%}
\newcommand{\NAll}{654}

\newcommand{\EthWhite}{60\%}
\newcommand{\EthBlack}{26\%}
\newcommand{\EthAsian}{6\%}
\newcommand{\EthMixed}{5\%}
\newcommand{\EthOther}{3\%}

\newcommand{\EtwoScaledChisq}{25.6}
\newcommand{\EtwoScaledDf}{4}
\newcommand{\EtwoScaledP}{<\,.001}

\newcommand{\ProbDeepgramH}{55\%}

\newcommand{\ProbDeepgramP}{40\%}

\newcommand{\ProbElevenLabsH}{51\%}

\newcommand{\ProbElevenLabsP}{61\%}

\newcommand{\ProbGoogleH}{33\%}

\newcommand{\ProbGoogleP}{49\%}

\newcommand{\ProbRimeH}{68\%}

\newcommand{\ProbRimeP}{49\%}

\newcommand{\ProbQDeepgram}{=\,.020}
\newcommand{\ProbQElevenLabs}{=\,.128}
\newcommand{\ProbQGoogle}{=\,.007}
\newcommand{\ProbQOpenAI}{=\,.583}
\newcommand{\ProbQRime}{=\,.004}

\newcommand{\SysOverActorH}{19\%}

\newcommand{\SysOverActorP}{36\%}

\newcommand{\SysOverActorCartesiaP}{49\%}
\newcommand{\SysOverActorCartesiaPCI}{42\%--57\%}

\newcommand{\VoiceRatioHCI}{0.56--1.04}
\newcommand{\VoiceRatioHBC}{0.82}

\newcommand{\VoiceSysLrtHP}{=\,.005}

\newcommand{\VoiceSDDeepgramH}{0.84}
\newcommand{\VoiceRatioPCI}{0.49--0.93}
\newcommand{\VoiceRatioPBC}{0.73}

\newcommand{\VoiceSysLrtPP}{=\,.031}

\newif\ifreserveresults
\reserveresultsfalse

\title{Human-Likeness and Conversational Preference Diverge\\ for Synthesized Agent Speech in Customer-Service Calls}
\name{\begin{tabular}{@{}c@{}}Jeffrey M. Girard, Antony D'Avirro, Jason Z. Zheng, \\ Jacqueline R. Vertino, and Benjamin Peloquin\end{tabular}}
\address{Fluid Concepts Research, San Francisco, USA}
\begin{document}
\maketitle
\begin{abstract}
Listeners can judge an agent's speech on how human it sounds (human-likeness) and on how much they would want to talk to it (conversational preference). Listening tests rarely ask about conversational preference, so the relation between the two judgments is unclear. We compared them in paired-choice tests on 120 turns that actors improvised as customer-service representatives. On each trial, two agents spoke the same turn. In Study~1, both agents came from six commercial text-to-speech systems, each with eight voices. In Study~2, one was the actor's original recording and the other a system. Some systems ranked higher on human-likeness, and others on conversational preference. Listeners chose systems over actors in \SysOverActorH{} of human-likeness trials but \SysOverActorP{} of conversational preference trials (and \SysOverActorCartesiaP{} for the most preferred system). On both judgments, variation among a system's voices was about three-quarters of that between systems. Of three automatic naturalness scores, two predicted lower human-likeness, and none predicted conversational preference. In this setting, listeners tended to favor faster speech and fewer pauses between clauses on both judgments, and more pitch variation for conversational preference. Thus, measuring only human-likeness or testing one voice per system can misjudge which agents listeners would rather talk to.
\end{abstract}
\section{Introduction}
\label{sec:intro}
Impressions of a customer-service representative (CSR) on a call are shaped by both what the CSR says and how they say it: their voice, pronunciation, and prosody (e.g., pacing, pausing, intonation, emphasis)~\cite{zhou2025serve}. Many voice agents are built as a cascade of separate components for speech recognition, a language model that writes the reply, and text-to-speech (TTS)~\cite{ji2024wavchat}. In such a cascade, the TTS component speaks the reply with little or no context about the conversation or the agent's role. Yet appropriate delivery, especially prosody, depends on context. The same words said with different prosody can signal different speaker intentions~\cite{hellbernd2016prosody} and alter perceptions of the speaker~\cite{voorveld2025persuasiveness,yu2024trustworthy}.

TTS quality is judged mainly through listening tests, which usually measure naturalness as human-likeness~\cite{cooper2024review}. This focus makes sense, because more human-sounding synthetic speech tends to be rated as more likable~\cite{baird2018likeability}. But Pandey et al.~\cite{pandey2025naturalness} argue that naturalness also includes how well the delivery fits the setting. For example, raters can expect a different delivery from CSRs than from audiobook narrators. A bored or curt CSR may sound more human than an enthusiastic one but still not be the one customers prefer. We therefore define \emph{conversational preference} as which agent a listener would rather talk to in a specific conversation, and we measure it separately from human-likeness.

Listening tests are also slow and costly, so automatic measures aim to stand in for them. Most predict ratings of naturalness or overall quality~\cite{cooper2024review}. Examples include the widely used UTMOS~\cite{saeki2022utmos}, its successor UTMOSv2~\cite{baba2024utmosv2}, and NISQA~\cite{mittag2020nisqatts}. Other measures focus on prosody. Some compare the distribution of prosodic features in synthetic speech with that in human recordings~\cite{minixhofer2026ttsds2}, and others score how likely the prosody is given the text~\cite{ulgen2026ttscore}. These prosody measures were validated mostly against human-likeness or naturalness. It is unclear whether naturalness scores or prosody measures track conversational preference, which may depend on other aspects of delivery. We therefore test whether UTMOS, UTMOSv2, and NISQA predict either judgment. Instead of overall prosody scores, we examine seven interpretable prosodic features, which point to adjustable aspects of delivery.

To examine how delivery, human-likeness, conversational preference, and automatic measures relate, we use the ToolTalk corpus~\cite{tooltalk_corpus}. Its trained actors improvised as customers and CSRs from written scenarios in an interactive tool-calling environment, where the CSRs used software tools to look up and update customer information. We study 120 of its CSR turns. We check each turn's transcript by hand and synthesize it with six commercial TTS systems. Each system offers many voices that can sound different, so we sample eight voices from each. Each CSR actor is randomly assigned one voice from each system, which speaks all of that actor's turns. In two paired-choice studies, listeners first hear the original recording of the customer's preceding turn and then two agents speaking the CSR's reply, with the same words but different delivery. Half the listeners judge which agent sounds more like a real person, and the other half which agent they would rather talk to. Both groups hear the same trials, so the two judgments differ only in what listeners are asked. In Study~1, the two agents come from different TTS systems. In Study~2, one agent is the CSR actor's original recording, and the other comes from one of the six systems.

Our study asks four questions. (1) Do human-likeness and conversational preference rank the TTS systems in the same order, and do the systems fall equally short of human speakers on both? (2) How much do voices within a system vary in listeners' judgments? (3) Do UTMOS, UTMOSv2, or NISQA predict either judgment? (4) Which interpretable prosodic features are associated with each judgment? The answers bear on four decisions in building and evaluating agents, one per question: which judgments to collect, how many voices to sample, how much to rely on automatic measures, and which aspects of delivery to tune and measure.

\ifreserveresults\clearpage\fi
\section{Methods}
\label{sec:methods}

\subsection{Speech Materials}
ToolTalk \cite{tooltalk_corpus} is a corpus of role-played customer-service calls in eight domains: airline, banking, food delivery, healthcare, hotel, restaurant, roadside assistance, and takeout.
We sampled 120 of its CSR turns of 8 to 30 words (median~14) at random, stratified by domain in proportion to the corpus. The sample included one or two turns from each of 97 actors. Parakeet TDT 0.6B v3 transcribed each turn, and we then checked the transcripts by hand.

Six TTS systems synthesized each turn from its transcript, without dialogue context or delivery instructions: Cartesia (Sonic 3.6), Deepgram (Aura-2), ElevenLabs (v4 Turbo), Google (Gemini 2.5 Flash TTS), OpenAI (gpt-4o-mini-tts), and Rime (Coda). We call each system's synthesis of a turn a \emph{render}. Each system contributed eight stock voices, half female. Within a system, all of a CSR actor's turns used one random voice of the actor's sex (38 female, 42 male), or of either sex if the actor's sex was unreported (17 actors).

From each original call, we extracted the caller's preceding turn as context. Study~2 also used the actor's recording of each turn. All audio passed through a simulated telephone channel (300--3,400 Hz band-pass, 8 kHz G.711 $\mu$-law, resampled to 16 kHz) and was set to a speech level of $-26$ dBFS. The channel was meant to mimic a phone call and reduce differences in audio quality between actors and systems; it does not alter timing and mostly preserves the pitch contour. We call each render or actor recording a \emph{clip}.

\begin{figure*}[t]
\centering
\includegraphics[width=\textwidth]{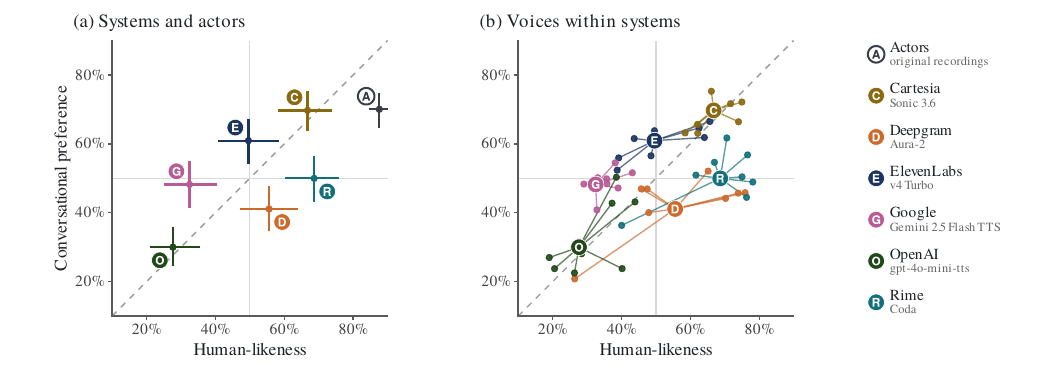}
\caption{Both axes give the probability of being chosen over an average system. (a) Systems and actors on the same scale from the baseline model fit to both studies (Section~\ref{sec:stats}), with 95\% CIs. Lettered circles label the points. (b) Each system's voices, with lines to its estimate from (a). Points above the diagonal do better on conversational preference than on human-likeness.}
\label{fig:scale}
\end{figure*}

\subsection{Listening Experiments}
Both experiments ran online in Gorilla, with listeners from Prolific. We followed the quality controls of the ITU-T P.808 standard for crowdsourced tests \cite{itu2021p808}. Eligible listeners lived in the US; used a computer, not a mobile device; spoke English natively; reported no hearing difficulties; and had an approval rate of at least 99\%. Headphones were recommended, not required, and listeners set a comfortable volume first.

Listeners were randomly assigned to one of two groups. One group answered ``Which agent sounds more like a real person?'' (human-likeness) on every trial, and the other answered ``Which agent would you prefer to talk to?'' (conversational preference). Each trial played the caller context and then the same turn spoken by Agent A and then by Agent B. Each comparison was played in both orders to different listeners, and listeners were not told whether an agent was human, or which system and voice it used. A session contained two practice trials, up to 30 main trials from 30 different turns in shuffled order, and four attention checks. In each attention check, one agent explicitly told the listener to choose it.

In Study~1, each trial compared the renders of the same turn from two different systems. Each turn appeared in six of the 15 possible system pairs, which gave 720 comparisons. Across turns, each pair of systems met 48~times. Each comparison was judged by nine to eleven listeners in each group. In Study~2, each trial compared the actor's recording of a turn with one of its six renders. Dropping two actor recordings (for audio issues) left 708 comparisons. Each was judged by four or five listeners in each group. The procedure was otherwise as in Study~1, and no listener could join both studies.

After quality control exclusions, Study~1 had \NB{} listeners (\NBHl{} human-likeness, \NBPr{} conversational preference) and \JudgmentsB{} judgments. Study~2 had \NBH{} listeners (\NBHHl{} human-likeness, \NBHPr{} conversational preference) and \JudgmentsBH{} judgments. Across both studies, listeners were aged \AgeMin--\AgeMax{} (mean \AgeMean, SD \AgeSD), and \FemaleAllN{} of \NAll{} (\FemaleAll) were female. By self-reported ethnicity, \EthWhite{} were White, \EthBlack{} Black, \EthAsian{} Asian, \EthMixed{} mixed, and \EthOther{} other or unreported.

\subsection{Automatic Measures}
\label{sec:compat}
UTMOS \cite{saeki2022utmos}, UTMOSv2 \cite{baba2024utmosv2}, and NISQA's naturalness model \cite{mittag2020nisqatts} are neural networks trained to predict listeners' naturalness ratings. UTMOSv2's score varies between runs, so we averaged five runs per clip. We applied all three to the clips as the listeners heard them, after the simulated telephone channel (as well as before it, as a sensitivity analysis).

We also measured seven interpretable prosodic features of delivery: the number of pauses of 0.2~s or longer between clauses (at punctuation) and within clauses (elsewhere), speaking rate (syllables per second of speech), phrase-final lengthening (phone duration before pauses versus elsewhere), pitch variation (SD of word pitch in semitones), final pitch slope (over the last word), and prominence variation (the SD across words of combined pitch, loudness, and duration). We extracted them with forced alignment \cite{mcauliffe2017mfa} and Praat via Parselmouth \cite{boersma2021praat,jadoul2018parselmouth} before the simulated telephone channel, because both are more accurate on full-band audio. 

\subsection{Statistical Analysis}
\label{sec:stats}
For each judgment, we modeled the Study~1 choices with a logistic mixed model, fit with glmmTMB~\cite{brooks2017glmmtmb}. This baseline model is a Bradley--Terry model~\cite{bradley1952rank} with random effects. Unlike raw win rates, it separates differences between systems from variation due to listeners, voices, and renders. When listener $\ell$ hears clip $a$ and then clip $b$ on trial $i$, it is
\begin{equation*}
 \mathrm{logit}\Pr(y_i{=}1)=\alpha_0+\alpha_\ell+\eta_a-\eta_b,
\end{equation*}
where $y_i=1$ if the listener chose clip $a$, and each clip $c$ has strength $\eta_c=\beta_{s(c)}+u_{v(c)}+w_c+g_{\ell s(c)}$. Here $\alpha_0$ is the overall A/B bias, and $\beta$ is the strength of each system. The random effects $\alpha$, $u$, $w$, and $g$ capture each listener's own A/B bias, the voice, the render, and the listener's view of each system.

To compare voices with systems, we divided the standard deviation of $u$ by that of the six system strengths. A parametric bootstrap with 2,000 simulated data sets gave this ratio a 95\% basic bootstrap confidence interval (CI) and an estimate of its bias, which we subtracted~\cite{davison1997bootstrap}. A likelihood-ratio test asked whether voice spread differs between systems.

A Wald test asked whether the conversational preference strengths were proportional to the human-likeness strengths, as they would be if listeners merely chose more consistently under one question. It treated the two judgments' fits as independent. For each system, we also estimated and tested how its probability of being chosen over an average system changed from human-likeness to conversational preference.

For Study~2, a logistic mixed model gave the probability of choosing each system over the actor. It had random effects for the listener, turn, voice, and render, and for each listener's own view of each system. We tested each probability against 0.5 (equal choice). CIs for these probabilities and for the Study~1 changes came from 4,000 draws from the fixed effects' sampling distribution. Study~1 probabilities are for a typical listener, voice, and render, and Study~2 probabilities average over listeners, turns, voices, renders, and play order. To place the actors on the systems' scale (Fig.~\ref{fig:scale}), we also fit the baseline model to both studies, with the actors as a seventh system and a random effect per actor.

We standardized each naturalness score and prosodic feature over the 720 Study~1 renders. We added the clips' difference in one naturalness score at a time to the baseline model, and we refit each model without the system strengths. With the system strengths, a score can explain only differences within systems; without them, it can also explain differences between systems. We then added the differences in all seven prosodic features instead of a naturalness score, again with and without the system strengths.\footnote{We tested the naturalness scores one at a time because each is meant to be used alone. We fit the prosodic features together so that each coefficient holds the others fixed; they were only weakly intercorrelated (all $|r|\le.20$).} Both clips spoke the same words, so these differences were matched for text. A second such test asked whether the feature coefficients differ between the judgments, overall and per feature. Within each set of related tests, we controlled the false discovery rate and report the adjusted $p$ values as $q$. Single tests keep unadjusted $p$.

\section{Results and Discussion}
\label{sec:results}

\subsection{The Two Judgments Rank Systems Differently}
In Study~1, the conversational preference strengths were not proportional to the human-likeness strengths ($\chi^2(\EtwoScaledDf)=\EtwoScaledChisq$, $p\EtwoScaledP$). OpenAI ranked lowest on both, and Cartesia ranked highest on conversational preference and nearly tied with Rime at the top on human-likeness. Neither moved significantly (both $q\ProbQOpenAI$). The other systems changed rank order (Fig.~\ref{fig:scale}a). From human-likeness to conversational preference, Rime's probability of being chosen over an average system fell from \ProbRimeH{} to \ProbRimeP{} ($q\ProbQRime$), and Deepgram's fell from \ProbDeepgramH{} to \ProbDeepgramP{} ($q\ProbQDeepgram$). Google's rose from \ProbGoogleH{} to \ProbGoogleP{} ($q\ProbQGoogle$), and ElevenLabs' rose from \ProbElevenLabsH{} to \ProbElevenLabsP{} ($q\ProbQElevenLabs$). Thus, an evaluation that collects only one of the two judgments can rank these systems differently.

\subsection{No System Matches the Actors on Human-Likeness}
\begin{figure}[t]
\centering
\includegraphics[width=\columnwidth]{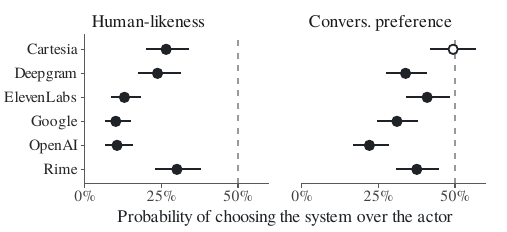}
\caption{Probability of choosing each system over the actor in Study~2. Bars show 95\% CIs, and filled points differ significantly from 50\% (equal choice, dashed).}
\label{fig:actor}
\end{figure}

In Study~2, every system's probability of being chosen over the actor for human-likeness was below 50\% (all $q<.001$, mean \SysOverActorH, Fig.~\ref{fig:actor}). For conversational preference, the mean rose to \SysOverActorP, but only Cartesia was chosen about as often as the actor (\SysOverActorCartesiaP{}, 95\% CI: \SysOverActorCartesiaPCI). Thus, a system that falls short of sounding human can still be preferred about as often as a person in the same role.

\subsection{Voices Within a System Differ Substantially}
After bias correction, the Study~1 ratio of the voice SD to the system SD was \VoiceRatioHBC{} (95\% CI: \VoiceRatioHCI) for human-likeness and \VoiceRatioPBC{} (\VoiceRatioPCI) for conversational preference. The voice SD also differed between systems ($p\VoiceSysLrtHP$ and $p\VoiceSysLrtPP$). On human-likeness, it ranged from below 0.01 for Google to \VoiceSDDeepgramH{} for Deepgram (logit scale, Fig.~\ref{fig:scale}b). Thus, both judgments depend on the voice as well as the system, and any evaluation with only one voice per system confounds the two.

\subsection{Higher UTMOS Predicts Lower Human-Likeness}
\begin{figure}[t]
\centering
\includegraphics[width=\columnwidth]{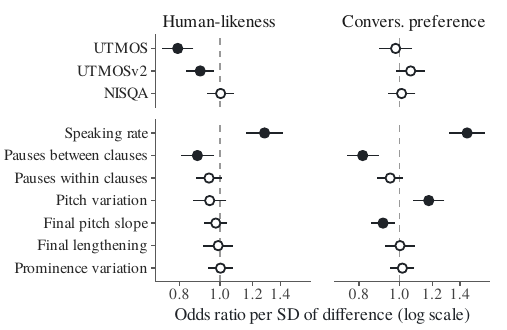}
\caption{Odds ratio for choosing a clip in Study~1, per SD of its advantage over the other clip. Scores (top) were fit separately and features (bottom) jointly. Bars show 95\% CIs, and filled points differ significantly from 1.0 (no association, dashed).}
\label{fig:measures}
\end{figure}

A clip with the higher UTMOS or UTMOSv2 score was less likely to be judged more human-like, and NISQA showed no association; none of the three reliably predicted conversational preference (Fig.~\ref{fig:measures}). These results held whether scores could explain only differences within systems or also differences between them. Listeners judged the actors the most human-like, but all three scores rated them below every system. The scores failed in the same ways when computed before the simulated telephone channel. Thus, no naturalness score could stand in for either judgment in this setting.

\subsection{Most Prosodic Associations Are Shared}
On both judgments, the clip with the faster speaking rate had higher odds of being chosen, and the clip with more pauses between clauses had lower odds (Fig.~\ref{fig:measures}). On conversational preference, the odds also rose with pitch variation and with a more falling final pitch slope. These associations held after false discovery rate control and were similar when the features could also explain differences between systems.

The feature coefficients differed overall between judgments ($\chi^2(7)=17.3$, $p=.016$), with only pitch variation surviving false discovery rate correction (ratio of odds ratios 1.25, 95\% CI: 1.10--1.41, $q=.005$). Thus, the two judgments share most of their prosodic associations, but only conversational preference favors higher pitch variation.

\subsection{Limitations and Conclusion}
We evaluated only the TTS component, and it spoke transcripts of human CSR turns rather than replies written by a language model, which can differ in wording and style. Our score and feature analyses are also correlational. Within these limits, our results suggest that (1)~human-likeness alone does not show which agent listeners would rather talk to, (2)~evaluations need many voices per system, modeled as random effects, (3)~naturalness scores need validation in each new setting, and (4)~speaking rate, pausing, and pitch variation are candidate targets when tuning delivery or designing measures.

\par\vfill\noindent\textbf{Acknowledgments.} Claude (Anthropic) helped write the analysis code and draft and edit all sections. The authors checked and take responsibility for all retained content.

\clearpage
\section{Compliance with Ethical Standards}
We judged that the study fits the US Common Rule exemption for benign behavioral studies with adults (45~CFR 46.104(d)(3)), so we did not seek external review. Listeners gave informed consent and were paid \$3.33 for an estimated 20 minutes. The ToolTalk actors agreed to research use of their recordings, including playback to new listeners.

\bibliographystyle{IEEEbib}
\bibliography{references}
\end{document}